\pdfoutput=1
\documentclass[11pt]{article}

\usepackage[]{emnlp2021}

\usepackage{times}
\usepackage{latexsym}

\usepackage{graphicx} % DO NOT CHANGE THIS
\usepackage{amsmath}
\usepackage{amssymb}
\usepackage{multicol,multirow}
\usepackage{tabularx,booktabs}       % professional-quality tables
\newcolumntype{Y}{>{\centering\arraybackslash}X}
\newcolumntype{s}{>{\hsize=.25\hsize}X}
\newcolumntype{m}{>{\hsize=.001\hsize}X}

\usepackage[T1]{fontenc}
\usepackage[utf8]{inputenc}

\usepackage{microtype}

\title{Leveraging Order-Free Tag Relations for Context-Aware Recommendation}
\author{Junmo Kang\qquad Jeonghwan Kim\qquad Suwon Shin\qquad Sung-Hyon Myaeng\\
  School of Computing, KAIST \\
  Daejeon, Republic of Korea \\
  {\tt \{junmo.kang, jeonghwankim123, ssw0093, myaeng\}@kaist.ac.kr} \\}

\begin{document}

\maketitle
\begin{abstract}
Tag recommendation relies on either a ranking function for top-$k$ tags or an autoregressive generation method. However, the previous methods neglect one of two seemingly conflicting yet desirable characteristics of a tag set: orderlessness and inter-dependency. While the ranking approach fails to address the inter-dependency among tags when they are ranked, the autoregressive approach fails to take orderlessness into account because it is designed to utilize sequential relations among tokens. We propose a \textit{sequence-oblivious} generation method for tag recommendation, in which the next tag to be generated is independent of the order of the generated tags and the order of the ground truth tags occurring in training data. Empirical results on two different domains, Instagram and Stack Overflow, show that our method is significantly superior to the previous approaches.
\end{abstract}

\section{Introduction}
\label{sec:intro}
\begin{figure}[t!]
    \centering
    \includegraphics[scale=0.5]{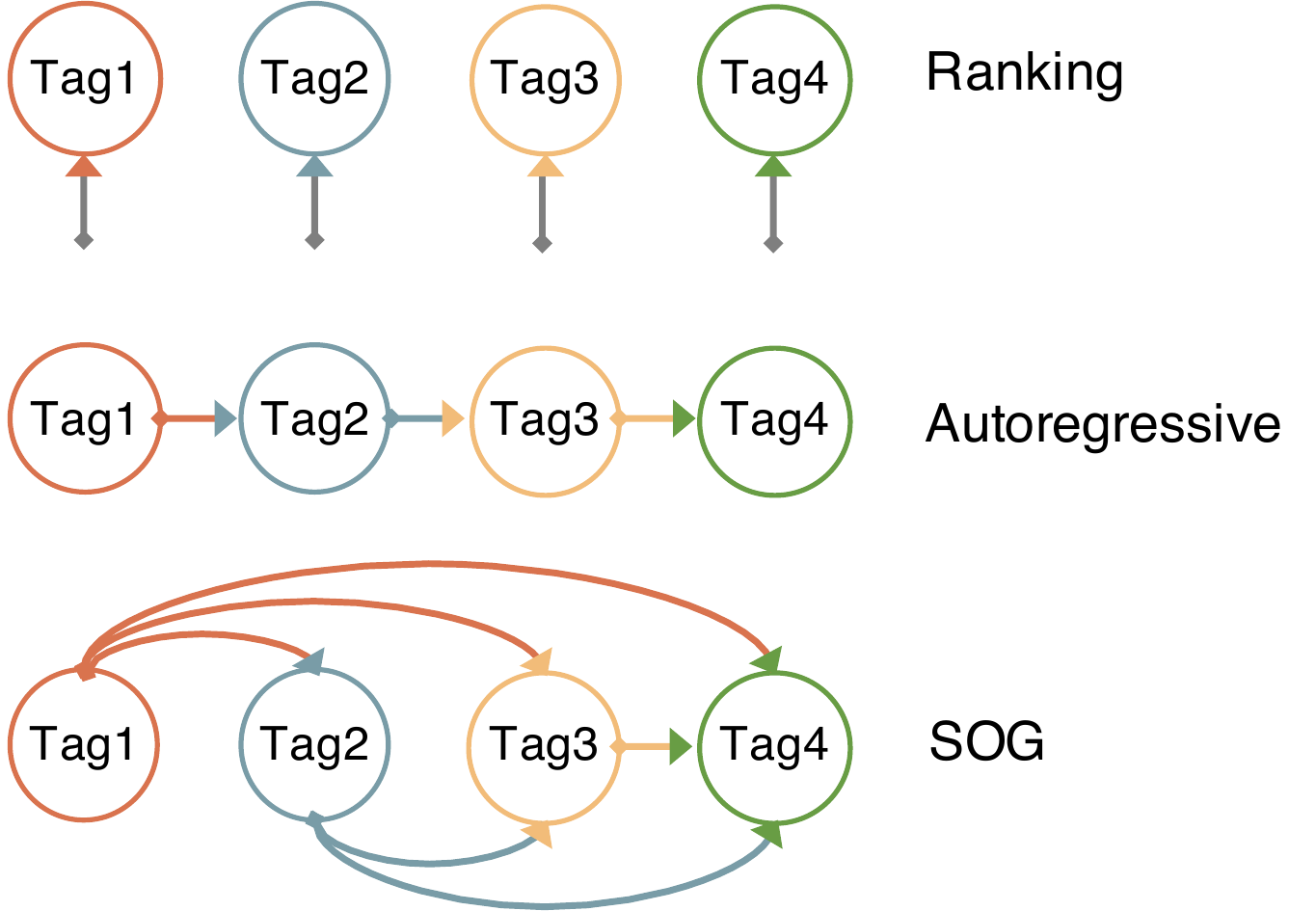}
    \caption{Graphical representation of three different frameworks for tag recommendation.}
    \label{fig:ranking_ar_sog}
\end{figure}

Recommendation techniques have been widely used for diverse applications both in symbolic and deep neural network frameworks. A conventional approach to recommendation is a ranking scheme that returns top-$k$ relevant target items for a given query or a user profile. As such, recommending tags (e.g., hashtags and labeled tags), which is the main focus of this paper, has been treated as a ranking problem \cite{weston-etal-2014-tagspace, gong2016hashtag, wu2018hashtag, wang2019microblog, zhang2019hashtag, yang2020sentiment, kaviani2020emhash}. These approaches, however, neglect \textit{inter-dependency} among the tags (see Figure \ref{fig:ranking_ar_sog})in a way conventional information retrieval techniques do for ranking. When tags have dependency among themselves, especially with a query, it would be desirable to consider such dependency when the next tag is selected for recommendation. 

On the opposite side of the spectrum are the recent studies employing an autoregressive (AR) generation model for tag recommendation \cite{wang2019microblog, yang2020amnn} where a GRU decoder \cite{cho-etal-2014-properties} enables the modeling of the dependency among the generated tags. While this approach considers dependency present in a sequence of tags, it overlooks the nature of tag recommendation where the output is a set of tags (\textit{orderlessness}) rather than a sequence. For tag recommendation, it is important that all the tags generated to a certain point in time have identical influence on the decision for the next generation (\textit{inter-dependency}) as in Figure \ref{fig:ranking_ar_sog}. 

Contrasting a list of tags against a sequence of words in a natural language sentence sheds light on the characteristics of the tag recommendation problem. Unlike an ordinary sentence comprised of a sequence of tokens, a set of recommended tags is not predicated upon any syntactic rule. It is unordered and syntax-free, and yet exhibits dependency among the tags themselves based on their semantic relatedness. These seemingly conflicting characteristics suggest that tag recommendation be seen as an application with a new class of language, calling for a new method beyond the text generation framework widely adopted for NLP \cite{cho-etal-2014-properties, bahdanau2014neural, sutskever2014sequence, radford2018improving}.

More specifically, we note that the decoder-oriented AR approach, even with a Transformer \cite{vaswani2017attention} decoder, is essentially limited in that the next token to be generated is highly dependent on the last generated token (see Figure \ref{fig:ranking_ar_sog}). While remembering what has been generated is helpful, furthermore, maximizing a sequence likelihood adopted in a typical text generation model is unnecessarily constraining for a tag recommendation; the \textit {order} of generated tags is immaterial as they could be shuffled without changing their overall relevance to the query.

% We propose our model here
In this paper, we propose the \textit{sequence-oblivious} generation (SOG) method for tag recommendation. Its main feature is that a tag is generated independently from the order of the previously generated tags. Instead, it iteratively expands the input query with the generated tag so that the expanded query affects the next tag prediction. %Moreover, it is designed to effectively fuse multi-modal input features with BERT \cite{Devlin2019BERTPO}.
Our approach can be seen as analogous to pseudo-relevance feedback (PRF) \cite{10.1145/243199.243202} where previously retrieved items are deemed relevant and used to extract additional query terms for relevance feedback. In our approach, previously generated tags are added to the query and fed back to the process of predicting the next tag to be generated.

Note that SOG is devised for the unique nature of generating syntax-free yet inter-dependent tags. Instead of the usual RNN or Transformer decoder where the sequence of tokens plays a key role, SOG leverages the Transformer \textit{encoder} and its self-attention to draw on the identical flow of information from every input feature position, removing the sequence-dependent nature of decoder-based generation. Moreover, our model is trained to ignore the order of the ground truth tags by maximizing on the entire set of tags (1-to-M) at every step of generation, which is distinguished from maximizing a sequence likelihood (1-to-1) in a typical autoregressive text generation task.

% Mention the fusion of multi-modal features (early vs. late)
Also proposed in this paper is a scheme for \textit{late} fusion of multi-modal context input, namely, text, image, time, location, and tags, as opposed to conventional \textit{late} fusion approaches.
%We also explore the effective fusion of multi-modal input for tag recommendation by evaluating the \textit{early} and conventional \textit{late} fusion approaches. 
We employ BERT \cite{Devlin2019BERTPO} for tag generation to exploit its self-attention mechanism for multi-modal fusion and its pre-trained language understanding capability. Our intuition is that input features should be encoded simultaneously for their mutual influences, not as separately encoded parts followed by their fusion at a later stage.

We conduct extensive experiments on two different domains (Instagram and Stack Overflow) for recommendation and show that SOG outperforms the ranking, AR generation and late fusion approaches by significant margins. We provide detailed analyses of the comparisons between SOG and established baselines to shed light on the different perspectives of the proposed approaches for recommendation.

\section{Related Work}
\label{sec:related_work}

The tag recommendation problem has mostly been studied as a ranking process, with models extracting top-$k$ tags given an input query \cite{zangerle2011recommending, weston-etal-2014-tagspace, sedhai2014hashtag, facebook_hashtag_prediction, park2016harrison, li2016hashtag, gong2016hashtag, Wu2018StarSpaceEA}. They map the input features and tag embeddings to a common embedding space and learn with a pairwise ranking loss \cite{weston-etal-2014-tagspace, facebook_hashtag_prediction, Wu2018StarSpaceEA}, or address tags as latent topics by applying topic modeling like Latent Dirichlet Allocation \cite{ding2012automatic,  godin2013using, zhao2016personalized, li2019topical}.

A recent line of work on tag recommendation takes a generative approach \cite{wang2019microblog, yang2020amnn}. However, their application of GRU shows that tags are treated as an ordered sequence, neglecting the orderless yet interrelated characteristics of tags. This characteristics of a tag set pose a significant challenge to the encoder-decoder generation scheme common to text generation \cite{cho-etal-2014-properties, bahdanau2014neural, sutskever2014sequence, radford2018improving, yang2019xlnet, chi2020cross}, where the autoregressive (AR) decoding mechanism heavily relies on the immediately preceding token.  This AR approach bounds the model to sequential ordering of the target sequence.

There have been attempts to use BERT in a generative fashion for text generation \cite{chan-fan-2019-bert, chan-fan-2019-recurrent}, exploiting its language understanding capabilities. While they use BERT to model sequential dependency, we explore its potential to ignore the sequential aspect. Recent work \cite{yang-etal-2019-deep} takes a reinforcement learning (RL) approach to modeling the unordered yet dependent characteristics of a target domain. Despite their learning objective to maximize reward on the orderless prediction, however, the model architecture still bound to the AR scheme.

\begin{figure*}[h!]
    \includegraphics[width=1.0\textwidth]{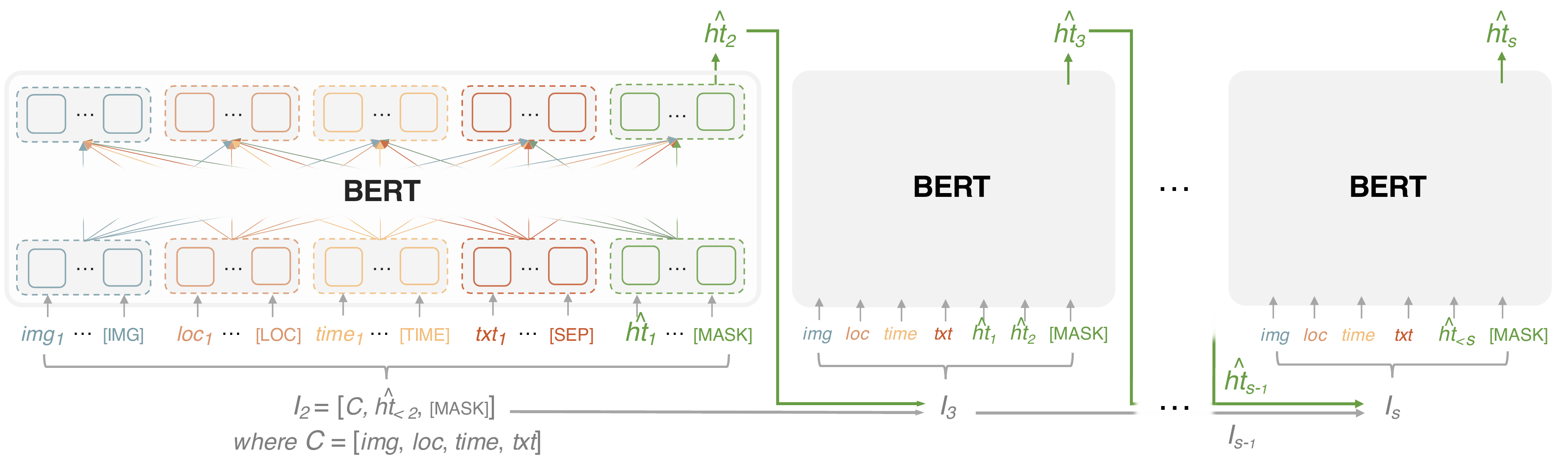}
    \caption{Overall architecture of the proposed model. It generates the most likely tag iteratively by leveraging BERT's self-attention and [MASK] token, allowing for an identical flow among generated tags.}
    \label{fig:architecture}
\end{figure*}

%\section{Sequence-Oblivious Generation}
\section{Context-Aware SOG Method}

Given the multi-modal context feature types such as image, location, time, and text as well as tags generated to a particular point in time, our recommendation model predicts the next tag using all of them as input query. More formally:

% {\small
\begin{align}
\hat{ht_s} = \operatorname*{argmax}_{ht_s \in {HT}} P(ht_{s} | img, loc, time, txt,\hat{ht}_{<s})
\label{eq:hashtag_generation}
\end{align}
% }
where $\hat{ht_s}$ is the $s^{th}$ tag to be generated by the model conditioned on the context features and previously generated unordered list of tags. $HT$ is the set of all possible tags.

The tag generation process is \textit{sequence-oblivious}, not following the way natural language sentence is generated. It does not abide by predefined syntactic or semantic rules and is oblivious to the sequential order. We define three indispensable components for SOG: (i) Encoder-based generation architecture with an unbiased exchange of information among tags (Section \ref{model}), (ii) 1-to-M training scheme that mitigates the order constraint (Section \ref{training}), and (iii) Greedy decoding that neglects the complex syntactic and semantic constraints of a natural language (Section \ref{decoding}).
Unlike the autoregressive generation or ranking schemes, our generation model produces an unordered yet interrelated target tags given an assortment of context features and previously generated tags. 
In Section \ref{fusion}, we elaborate on how the four context feature types are fused for our SOG model.
% Our model employs a pre-trained BERT, which is fine-tuned to generate tags co-referenced with the context features (i.e., image, location, time and text) of a given post instance as in Figure \ref{fig:architecture}. 
%We propose the \textit{Sequence-Oblivious} approach to defining the (i) model structure, (ii) training, and (iii) decoding steps.

% Our sequence-oblivious approach not only proposes a novel generation scheme, but also provides an alternative generation method for recommendation models that require the modeling of syntax-free and orderless nature of the target domain. 

\subsection{Sequence-Oblivious Model} \label{model}
As a part of establishing the sequence-oblivious characteristic, Transformer \cite{vaswani2017attention} encoder, not decoder, is adopted as an architecture for generation.
Specifically, our approach exploits BERT \cite{Devlin2019BERTPO}, a pre-trained language model based on Transformer encoder that is capable of bidirectional encoding.
% language model that takes as input a sequence of partially masked words and predicts which word should originally be placed in the \texttt{[MASK]} token's position.
While it is not originally devised for generation purpose, we note on: (i) the ability of a \texttt{[MASK]} token to aggregate surrounding contextual information and (ii) the Transformer encoder's characteristic to mitigate sequential elements of a given input through bidirectionality and enable identical flow of information via self-attention. 
We remove original positional encoding and utilize the context-aggregating characteristic of \texttt{[MASK]} tokens to compute a probability distribution over the set of target tags. This \textit{encoder} style generation can be contrasted with an autoregressive \textit{decoder} style generation. While the latter generates a token at a given timestep $t$ by only pooling from the token at $t-1$, the former is structurally free from such a constraint. 

%This is especially true when dealing with sequence-independent problems like tag recommendation.

Given a sequence of $n$ tokens  $x = [x_{1}, x_{2}, x_{3}, ..., x_{n}]$ representing the initial input context (the features in Eq.\ref{eq:hashtag_generation}) to the encoder, our model begins its generation process with:

% {\small
\begin{align}
    X_{1} &= [x, \texttt{[SEP]}, \texttt{[MASK]}]
\end{align}
% }
where \texttt{[SEP]} indicates the end of the input sequence, and \texttt{[MASK]} is appended at the end as a \textit{generative} token over a target vocabulary set $V$ (henceforth $HT$ for the tag set in this work). As the model generates one token after another, the predicted tokens are consecutively accumulated to expand the input context to be fed back to the model:

% {\small
\begin{align}
    X_{i} &= [x, \texttt{[SEP]}, \hat{y}_{1}, \hat{y}_{2}, ..., \hat{y}_{i-1}, \texttt{[MASK]}]
\end{align}
% }
where $X_{i}$ refers to the $i^{th}$ input of a given data instance.
% Following the concatenation of the generated sequence of tokens between the \texttt{[SEP]} and \texttt{[MASK]} tokens, the model feeds it back into itself and takes the representation from the \texttt{[MASK]} token's position to generate the subsequent token:
Following the construction of the input $X_i$, the model feeds it back into itself and takes the \texttt{[MASK]} token representation aggregating all information of context and tags to generate the subsequent tag:

% {\small
% \begin{align}
%     \boldsymbol{h_{i}} &= BERT(X_{i}) \label{eq:bert_hidden} \\
%     \boldsymbol{z} &= \boldsymbol{h}_{i}^{\texttt{[MASK]}} \cdot \boldsymbol{W}_{vocab}^T + b \\
%     p_{k} &= \frac{\exp(\boldsymbol{z}_{k})}{\sum_{ht \in HT} \exp(\boldsymbol{z}_{ht})} \label{eq:bert_dist} \\
%     \hat{y}_{i} &= \operatorname*{argmax}_{y_{i}}(\boldsymbol{p}) \label{eq:bert_argmax}
% \end{align}
% }
% {\small
\begin{align}
    \boldsymbol{h_{i}} &= BERT(X_{i}) \\
    \boldsymbol{z} &= \boldsymbol{h}_{i}^{\texttt{[MASK]}} \cdot \boldsymbol{W}_{HT}^T + b \\
    p_{k} &= \frac{\exp(\boldsymbol{z}_{k})}{\sum_{ht \in HT} \exp(\boldsymbol{z}_{ht})} \\ 
    \hat{y}_{i} &= \operatorname*{argmax}_{y_{i}}(\boldsymbol{p})
\end{align}
% }
% {\small
% \begin{align}
%     \boldsymbol{z} &= \boldsymbol{h}_{i}^{\texttt{[MASK]}} \cdot \boldsymbol{W}_{HT}^T + b
% \end{align}
% }
% {\small
% \begin{align}
%     p_{k} &= \frac{\exp(\boldsymbol{z}_{k})}{\sum_{ht \in HT} \exp(\boldsymbol{z}_{ht})} \label{eq:bert_dist}
% \end{align}
% }    
% {\small
% \begin{align}
%     \hat{y}_{i} &= \operatorname*{argmax}_{y_{i}}(\boldsymbol{p}) \label{eq:bert_argmax}
% \end{align}
% }
where $\boldsymbol{h}_{i} \in \mathbb{R}^{n \times d}$, $\boldsymbol{W}_{HT} \in \mathbb{R}^{|HT| \times d}$, $\boldsymbol{z} \in \mathbb{R}^{|HT|}$, and $\boldsymbol{p} \in \mathbb{R}^{|HT|}$. Then the generated $\hat{y}_{i}$ is again used to expand the input sequence for the next input:

% {\small
\begin{align}
    X_{i+1} = [x, \texttt{[SEP]}, \hat{y}_{1}, \hat{y}_{2} ..., \hat{y}_{i-1}, \hat{y}_{i}, \texttt{[MASK]}]
\end{align}
% }

\subsection{Sequence-Oblivious Training} \label{training}
\begin{figure}[h!]
    \centering
    \includegraphics[scale=0.48]{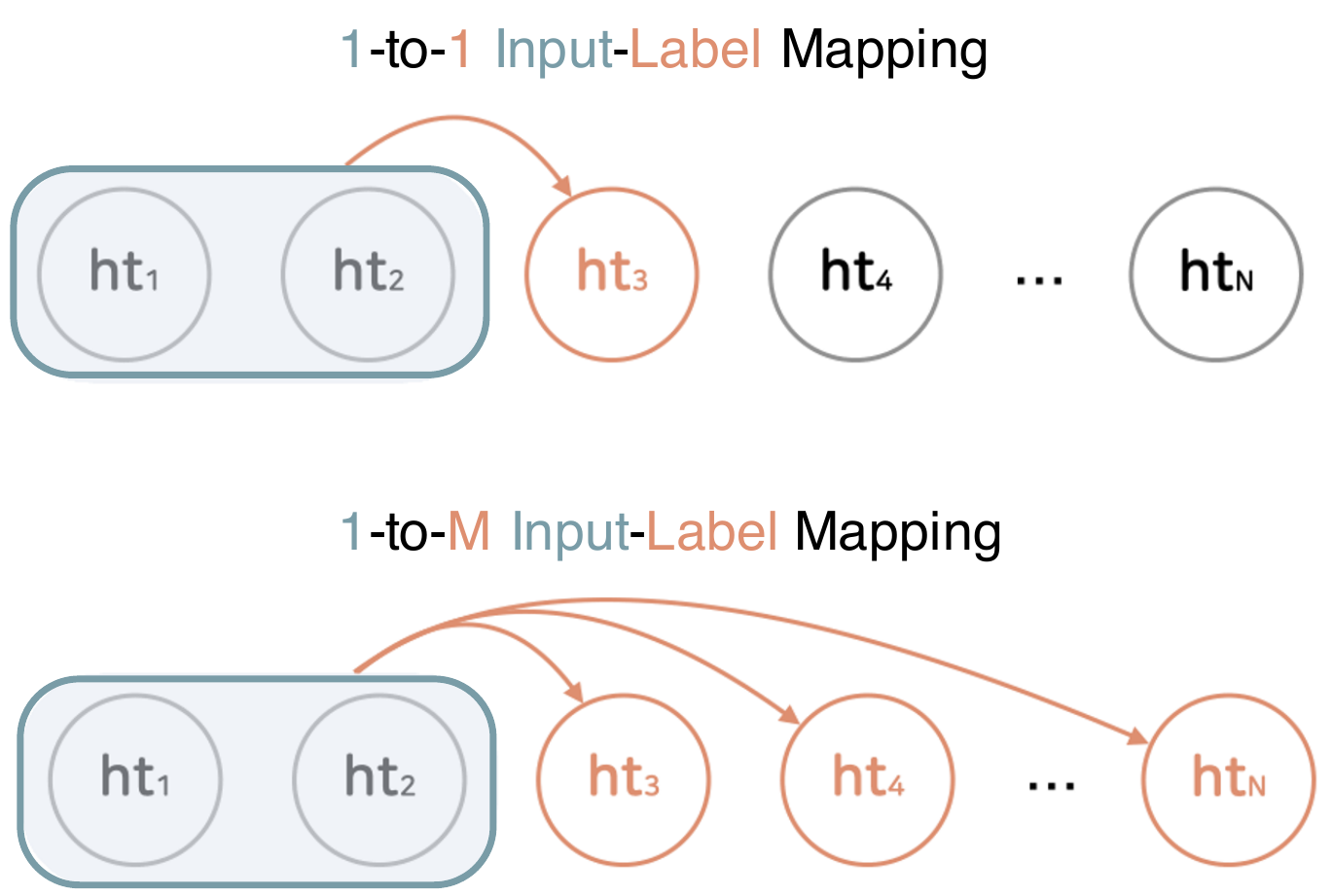}
    \caption{1-to-1 vs. 1-to-M training schemes. The hashtag ht{\tiny 3} that appears right after ht{\tiny 2} is not necessarily the ground truth for training; all permutations of the tags are considered ground truths. Moreover, unlike AR that is highly dependent on the last tag ht{\tiny 2}, our sequence-oblivious model allows for the equal contribution of ht{\tiny 1} and ht{\tiny 2} when generating ht{\tiny 3} to ht{\tiny N}.}
    \label{fig:1-to-1_vs._1-to-M}
\end{figure}
Training for conventional text generation usually consists of a single ground truth label at every generation step, which we refer to as the \textit{1-to-1} scheme. Such a method is reasonable since the exact sequence needs to be generated. In tag recommendation, however, multiple ground truth tags can exist that range over all the permutations of the given set of tags. This is what we refer to as the \textit{1-to-M} scheme (Figure \ref{fig:1-to-1_vs._1-to-M}). 

More formally, 
$L_{i}$, the ground truth tag(s) at the $i^{th}$ step, is defined in two ways:

% {\small
\begin{align}
    &L^{\{1-to-1\}}_{i} = ht_{i}\\
    &L^{\{1-to-M\}}_{i} = [ht_{i}, ht_{i+1}, ..., ht_{N}]
\end{align}
% }
where $ht_{i}$ is the $i^{th}$ ground truth tag. Note that the list of tags in $L^{\{1-to-M\}}_{i}$ can be any one of the permutations of those in the target tag set. We enforce the $L^{\{1-to-M\}}_{i}$ relationship by using KL divergence loss %(Eq. \ref{eq:kl-divergence}) 
that compares the output distribution of predicted tags against the ground truth distribution.

%We use the 1-to-M training scheme since the order by which the ground truth tags are generated does not matter. %For example, the next tag that a model needs to generate is not necessarily $ht_2$ when $ht_1$ is part of the input, even if they appear in that order in the post. 

% {\small
\begin{equation} \label{eq:kl-divergence}
D_{KL}(\boldsymbol{q}||\boldsymbol{p}) = \sum_{ht \in HT} \boldsymbol{q}(ht) \log \frac{\boldsymbol{q}(ht)}{\boldsymbol{p}(ht)}
\end{equation}
% }
where $ht$ refers to a tag within tag space of $HT$.

We induce inter-tag relationship at the feature level by turning a post containing $N$ ground truth tags into $N$ separate training instances. 
Each instance begins with the context features [$img$, $loc$, $time$, $txt$] and receives no tag, one tag, and so on, all the way to the maximum number of tags minus one. Starting with the first training instance $T_{1}$ being [$C$, \texttt{[SEP]}, \texttt{[MASK]}] where $C$ denotes the list of context features, we obtain a total of $N$ training instances:

% {\small
\begin{align} \label{eq:training_instance}
      T_i &= [C, \texttt{[SEP]}, ht_1, ht_2, ..., ht_{i-1}, \texttt{[MASK]}]
\end{align}
% }
where $i= 1,..., N$.

\subsection{Sequence-Oblivious Decoding Strategy} \label{decoding}
At the decoding step where a list of relevant tags is recommended in a sequence-oblivious way, our framework employs \textit{greedy search}. We avoid generating tags already generated by setting their probabilities to 0.
In text generation, beam search is regarded as a default setting since making a sentence natural and understandable is important. However, recommendation tasks prefer greedy search because it only cares for choosing the most relevant tags at each step of generation, without having to consider all the possibilities to satisfy the complex syntactic and semantic constraints. %For instance, given two lists of tag recommendation with one suggesting the most probable item at the $1^{st}$ step and the other suggesting it at the $10^{th}$ step of generation, the prior is clearly determined to be a better alternative. 

If beam search is used, there is a possibility of selecting the most likely ``sequence," not the list of items enumerated by their relevance to a given query, leaving out more relevant ones.
Hence, generating the most probable item at each step is more suitable when applying the generation framework to recommendation tasks. The sequence-oblivious and relevance-maximizing characteristics of this decoding strategy reinforces our concept of SOG. %in the light of recommendation.

\subsection{Feature Conversion and Early Fusion} \label{fusion}
In dealing with multi-modal context, we convert the features of four types\footnote{These are used only in the Instagram dataset, one of the two used for our experiments.} into their textual forms to make the different modalities amenable to BERT; their weighted representations are fused through the self-attention mechanism. The pre-processing of each feature is as follows:

\textbf{Image:} We generate a caption from each image using the Microsoft Azure image captioning module. Note that while any module can be adopted, we chose it because it is a commercialized tool, which can be verified easily and reliably by others. Only the first image in each post is used.

\textbf{Location:} We utilize the symbolic names (e.g., My Home, XX National Park) given by users. %ignoring coordinate information such as GPS location.

\textbf{Time:} A numeric time expression is converted into words in three categories based on a rule-based converter: season, day of the week, and part of the day (i.e. morning, afternoon, evening or night).  For example, '2020-07-01 (Wed) 14:52:00' is converted into \{summer, weekday, afternoon\}.

\textbf{Text:} A list of words are collected from the text description of a post. We strip tags from the description and use texts only.

We then enter the input context $C$ of $img$, $loc$, $time$ and $txt$ with a delimiter token for each type (\texttt{[IMG]}, \texttt{[LOC]}, \texttt{[TIME]}, \texttt{[SEP]}). Note that the tag $\hat{ht}$ generated from the previous stage and \texttt{[MASK]} token is appended at the end:

{\small
\begin{align}
    img &= [img_1, img_2, ..., img_{|img|-1}, \texttt{[IMG]}] \\
    loc &= [loc_1, loc_2, ..., loc_{|loc|-1}, \texttt{[LOC]}] \\
    time &= [time_1, time_2, ..., time_{|time|-1}, \texttt{[TIME]}] \\
    txt &= [txt_1, txt_2, ..., txt_{|txt|-1}, \texttt{[SEP]}] \\
    C &= [img, loc, time, txt] \label{eq:context_input}\\
    I_{s} &= [C, \hat{ht}_{<s}, \texttt{[MASK]}] \label{eq:input_seq}
\end{align}
}
where $I_{s}$ is the input at $s^{th}$ step for a given post.

Following feature conversion, SOG employs the \textit{early fusion} of input features with self-attention, which effectively merges the features representing multiple modalities that are interrelated and mutually complementary to one another. For example, two different representations can be constructed for an image of the sun along with time features given as either ``morning" or ``night," which will generate either ``sun rise" or ``sun set" as the target tag, respectively. As such, fusing inter-contextual information at the representation construction step is crucial for generating the relevant tags.

\section{Experiments}

\subsection{Experimental Settings}

\subsubsection{Datasets} \label{subsubsec:Datasets}
\begin{table}[h!]
    {\small
    \centering
    \begin{tabularx}{\linewidth}{ c *{5}{Y} }
    \toprule
        \multirow{2}{*}{\textbf{Dataset}} & \multirow{2}{*}{\textbf{\#Posts}} & \multirow{2}{*}{\textbf{\#Tags}} & \textbf{\#Tags / Post} & \multirow{2}{*}{\textbf{Input}} \\ \midrule\midrule
        \textbf{Instagram} & 87,872 & 907 & 9.59 & All \\ 
        \textbf{Stack Overflow} & 81,320 & 3,897 & 3.05 & Text \\
        \bottomrule
    \end{tabularx}
    }
    \caption{Two datasets used. The train/test ratio is 9:1. ``All" refers to all the context features in \textbf{Instagram}: Image, Location, Time, and Text. The "Text" of \textbf{Stack Overflow} means the title and the body of questions.}
    \label{table:dataset_statistics}
\end{table}
We evaluate our method on two different domains: Instagram and Stack Overflow. For the Instagram dataset, which is our main benchmark, we use multi-modal contexts to demonstrate their usefulness as well as the effectiveness of our fusion method through a comprehensive set of experiments. While Instagram has been adopted as a dataset in the previous work \cite{park2016harrison}, it only has a single modality (i.e., image) and the context features required in our approach are absent, so we built our own dataset.
For the Stack Overflow dataset, we use the text modality only, because context information such as time in Stack Overflow are not associated with predicting correct tag set. Despite the use of only the single modality, the dataset is further employed to evaluate the SOG method with additional experiments for generalizability. Overall statistics are in Table \ref{table:dataset_statistics} while details on the data collection and the overall process are in the Appendix.

\textbf{Instagram} is a popular photo sharing social network service (SNS). A post on Instagram contains images, location information, uploading time, text description and the corresponding hashtags (i.e., tags). We refer to the first four features as the \textit{context} and pair it with the associated tags.

\textbf{Stack Overflow} is a programming Q \& A community where each post contains a question, a list of answers, and user-annotated tags that summarize the topic at hand (e.g., java, nlp, pytorch). We use the questions as our input and tags as our target labels. Among the attributes (title, body, date, reputation scorem and user ID) each question contains, we only use \textit{title} and \textit{body} as our input text feature, excluding the others as they are rarely relevant. For example, unlike in Instagram, the time of a day is less likely to determine the contents of a question (and tags).

We consider these datasets as our benchmarks, containing a relatively sufficient number of tags (i.e., classes), to conduct a valid assessment of the effectiveness of leveraging the nature of tag relations. Compared to these datasets, other existing datasets for multi-label classification, contain only a few classes (e.g., 103 for RCV1-V2 \cite{10.5555/1005332.1005345} and 54 for AAPD \cite{yang-etal-2018-sgm}), as well as longer text that is richer in information. This forms a strong and direct association between context and a set of tags, which can obscure the effect of our method of iteratively adding information with tags for building sufficient context information.

\begingroup
\begin{table*}[h]
\centering
% {\small
\begin{tabularx}{\textwidth}{ Yssssssmssssss }
\toprule
 & \multicolumn{6}{c}{\normalsize{\textbf{Instagram}}} & &  \multicolumn{6}{c}{\normalsize{\textbf{Stack Overflow}}} \\
\normalsize{\textbf{Model}} & \small{\textbf{P@1}} & \small{\textbf{P@3}} & \small{\textbf{P@5}} & \small{\textbf{R@1}} & \small{\textbf{R@3}} & \small{\textbf{R@5}} & & \small{\textbf{P@1}} & \small{\textbf{P@3}} & \small{\textbf{P@5}} & \small{\textbf{R@1}} & \small{\textbf{R@3}} & \small{\textbf{R@5}} \\ 
\midrule\midrule
\multicolumn{14}{c}{\normalsize{\textbf{Ranking Approaches}}} \\
\midrule
\small{Frequency-Based}   &   6.99   &   5.33   &   4.65   &   1.56   &   4.22   &   5.76   &   &   7.49   &   8.06   &   7.57   &   3.07   &   10.36   &   16.73   \\  
\small{Joint Space}  &   15.21   &   11.30   &   9.58   &   6.21   &   12.78   &   17.40   &   &   9.70   &   7.30   &   6.06   &   4.31   &   9.64   &   13.26   \\ 
\small{BR-LF}  &  28.84  &  19.05  &  14.88  &  11.34  &  21.24  &  26.77  &  &  -   &   -   &   -   &   -   &   -   &   -   \\
\small{BR-EF}  &   41.78   &   26.52   &   20.51   &   17.52   &   28.69   &   34.50   &   &   58.72   &   30.07   &   20.56   &   27.93   &   39.69   &   44.01   \\
\midrule
\multicolumn{14}{c}{\normalsize{\textbf{Generation Approaches}}} \\
\midrule
\small{Seq2Seq (MLE)}  &   34.05   &   18.94   &   13.24   &   13.87   &   20.09   &   22.30   &   &   53.38   &   25.93   &   16.44   &   25.32   &   34.40   &   35.83   \\
\small{Seq2Set (MLE+RL)}  &   39.96   &   21.49   &   14.66   &   16.77   &   24.95   &   26.94   &   &   57.30   &   26.72   &   16.95   &   27.36   &   36.00   &   37.53   \\
\small{AR (1-to-1)}  &   30.06   &   18.54   &   14.66   &   12.83   &   20.49   &   25.88   &   &   61.66   &   31.46   &   21.06   &   29.24   &   42.07   &   45.98   \\
\small{AR (1-to-M)}  &   37.34   &   22.33   &   17.46   &   15.53   &   24.04   &   29.47   &   &   62.30   &   32.29   &   22.15   &   29.64   &   42.76   &   47.49   \\
\midrule
\small{\textbf{SOG (1-to-1)}}   &   46.15   &   28.03   &   20.58   &   19.81   &   30.44   &   34.65   &   &   59.80   &   34.30   &   \textbf{24.23}   &   29.20   &   46.81   &   \textbf{52.84}   \\ 
\small{\textbf{SOG (1-to-M)}}   &   \textbf{51.85}   &   \textbf{30.15}   &   \textbf{22.03}   &   \textbf{22.29}   &   \textbf{32.38}   &   \textbf{36.62}   &   &   \textbf{67.87}   &   \textbf{35.91}   &   24.09   &   \textbf{32.69}   &   \textbf{47.51}   &   51.82   \\ \bottomrule
\end{tabularx}
% }
\caption{Evaluation results of baselines (Ranking and Autoregressive) and our models on two different domains.}
\label{table:evaluation}
\end{table*}
\endgroup

\subsubsection{Metrics}
We evaluate with precision-at-k (P@K) and recall-at-k (R@K), with K being 1, 3 and 5. Both are widely used for recommendation tasks since the rank of the correctly predicted tags matters, irrespective of the order of the ground truth tags.

% {\small
\begin{align*}
P@K &= \frac{1}{N} \sum \frac{|\textrm{Ranked top-$K$} \cap {\textrm{Ground Truth}}|}{K}
\\
R@K &= \frac{1}{N} \sum \frac{|\textrm{Ranked top-$K$} \cap \textrm{Ground Truth}|}{|\textrm{Ground Truth}|}
\end{align*}
% }

\subsubsection{Baselines}
We compare our model against both the (i) ranking and (ii) autoregressive frameworks. For fair comparisons, we design the representative versions of the baselines that contain the core properties of each framework by disregarding task specific techniques (e.g., leveraging user metadata). Note that all the baselines and our model use the same context input features. We evaluate the following baselines under our settings:

\textbf{Frequency-Based:} A simple baseline to establish a lower bound by generating most frequent tags regardless of a given context. This shows the impact of frequency bias on the constructed data set.

\textbf{Joint Space (Ranking):} A generalized version of conventional tag recommendation models using the top-$k$ ranking framework \cite{weston-etal-2014-tagspace, facebook_hashtag_prediction, Wu2018StarSpaceEA, yang2020sentiment}. It projects input and tag embeddings onto the same representation space and learn with a pairwise ranking loss.

\textbf{BERT-based Ranking (BR-EF vs. BR-LF):} A modified version of the \textbf{Joint Space (Ranking)} baseline, using BERT as the backbone architecture. This model is trained with cross-entropy to maximize the likelihood of ground truth tags, and produces top-$k$ tags given the \texttt{[CLS]} representation. We compare two models, BR-EF and BR-LFT, for our early fusion (EF) and late fusion (LF) taken by previous models \cite{weston-etal-2014-tagspace, gong2016hashtag, wu2018hashtag, zhang2019hashtag, yang2020sentiment, kaviani2020emhash}, respectively. BR-EF takes the input $\textit{C}$ (Eq. \ref{eq:context_input}) whereas BR-LF passes each feature separately through BERT to independently encode each feature for the subsequent fusion step.

\textbf{Seq2Seq (MLE):}
A model used in \cite{yang-etal-2018-sgm, yang-etal-2019-deep} that employs the Seq2Seq framework \cite{sutskever2014sequence} for multi-label classification and learns via maximum likelihood estimation (MLE). The architecture consists of a bi-LSTM encoder and an LSTM decoder.

\textbf{Seq2Set (MLE+RL):}
A model \cite{yang-etal-2019-deep} built upon \textbf{Seq2Seq (MLE)}, using the Seq2Seq model pre-trained with MLE and fine-tuned with reinforcement learning (RL) to reduce the sensitivity to the label order in Seq2Seq through an F1-score based reward function.

\textbf{AR (1-to-1 vs. 1-to-M):} 
A generalized version of the autoregressive (AR) tag generation models \cite{wang2019microblog, yang2020amnn}. We employ the Transformer \cite{vaswani2017attention} encoder-decoder architecture, but replace the encoder with BERT for fair comparison with our SOG. We also apply the 1-to-M scheme to validate the effect.

\subsubsection{Implementation details}
Our implementations are based on bert-base-uncased of the \texttt{transformers} library \cite{wolf2019transformers}, using a V100 NVIDIA GPU. With the batch size of 64, the hidden size of 768 and the learning rate of 5e-5, we use the Adam optimizer and a seed equal to 42. The maximum input sequence length of our model is set to 384. The number of tags to be generated is set to 5 for our experiments, while it could be any number according to the use case.
Further details are in the Appendix.
% \ref{hyperparameters}

\subsection{Analyses of the Comparisons}
 Table \ref{table:evaluation} presents P@K and R@K scores of our models against those of the baselines, showing that our models outperform the baselines by significant margins. The most salient outcome of the experiment is that our proposed model is much superior to the ranking and autoregressive generation approaches. Another notable result is that the EF strategy is superior to LF. Further analyses follow:

\textbf{Joint Space vs. BR-LF:}
This comparison implies that the language model capabilities of BERT contributes significantly to the tag recommendation problem, compared to the joint space approach.

\textbf{BR-EF vs. BR-LF:} We assess the EF and LF approaches for inter-context feature modeling by comparing BR-EF and BR-LF. 
In LF, the model separately encodes each feature type (image, location, time and text) with a shared parameter BERT and averages over them to form a single, aggregated context representation. On the other hand, EF jointly feeds the features of different types at a single step and rank the tags based on the fused information. The large gap in the result clearly indicates that fusing the different features, rather than just aggregating them, helps the recommendation tasks significantly.  %As the results between these two models imply, EF provides an extensive and more comprehensive view of the given features unlike the delayed aggregation of separately modeled context representations.

\textbf{Ranking vs. Generation:} 
The result shows that SOG outperforms the ranking models by a significant margin. This performance gain is attributed to the modeling of the orderless inter-tag dependency. Note, however, that comparison between the ranking models and the AR models is mixed; the former is better for Instagram but worse for Stack Overflow. 

%In contrast to SOG's performance, AR models find it difficult to beat the ranking models as in Table \ref{table:evaluation}. Although the AR models appear to marginally outperform the ranking counterparts for Stack Overflow, the performance drop is substantial when tested on the Instagram data set. This result can be attributed to the sequential nature of the AR models, where the models are susceptible to the direct influence of the immediately preceding tag. %Such property of AR models refrain them from effectively utilizing the information flow from every input feature equally, leading to lower performance than the ranking model.

\textbf{AR vs. SOG:}
SOG models substantially outperform the AR models, supporting the claim that it is important not to emphasize the usual sequential dependency enforced by the AR models for tag recommendation.
%because the former ensures the identical flow of information with the independent and identical aggregation of input features by [MASK] token whereas the latter relies on the immediately preceding tag to generate the next one.
Our model also outperforms both Seq2Set (MLE+RL) and its backbone model Seq2Seq (MLE). Despite Seq2Set's improvement over Seq2Seq, it still underperforms SOG by a large margin. These results imply the architectural limitations imposed by the Seq2Seq backbone with the usual LSTM encoder-decoder; this structure ultimately confines the model to the AR framework. Moreover, the RL algorithms' notoriously poor sample efficiency \cite{Clark2020ELECTRA:} under the large action space of the entire tag set limits the performance of Seq2Set. Another limitation is that MLE pre-training is prerequisite for RL training.

\textbf{1-to-1 vs. 1-to-M:} For both the AR model and our model, we observe the 1-to-M models outperform 1-to-1 by a significant gap, which shows the effectiveness of our approach under the \textit{orderlessness} assumption. Note that the number of training instances of 1-to-1 and 1-to-M are exactly the same, meaning that the improvement is not due to data augmentation but to the advantage of mitigating the ordering constraint.
To test orderlessness, in addition, we shuffle the order of tags within the posts and train our model under the same setting. 
There is no meaningful gap in performance with the original result, validating our assumption.

\begin{figure}[t]
    \includegraphics[width=0.48\textwidth]{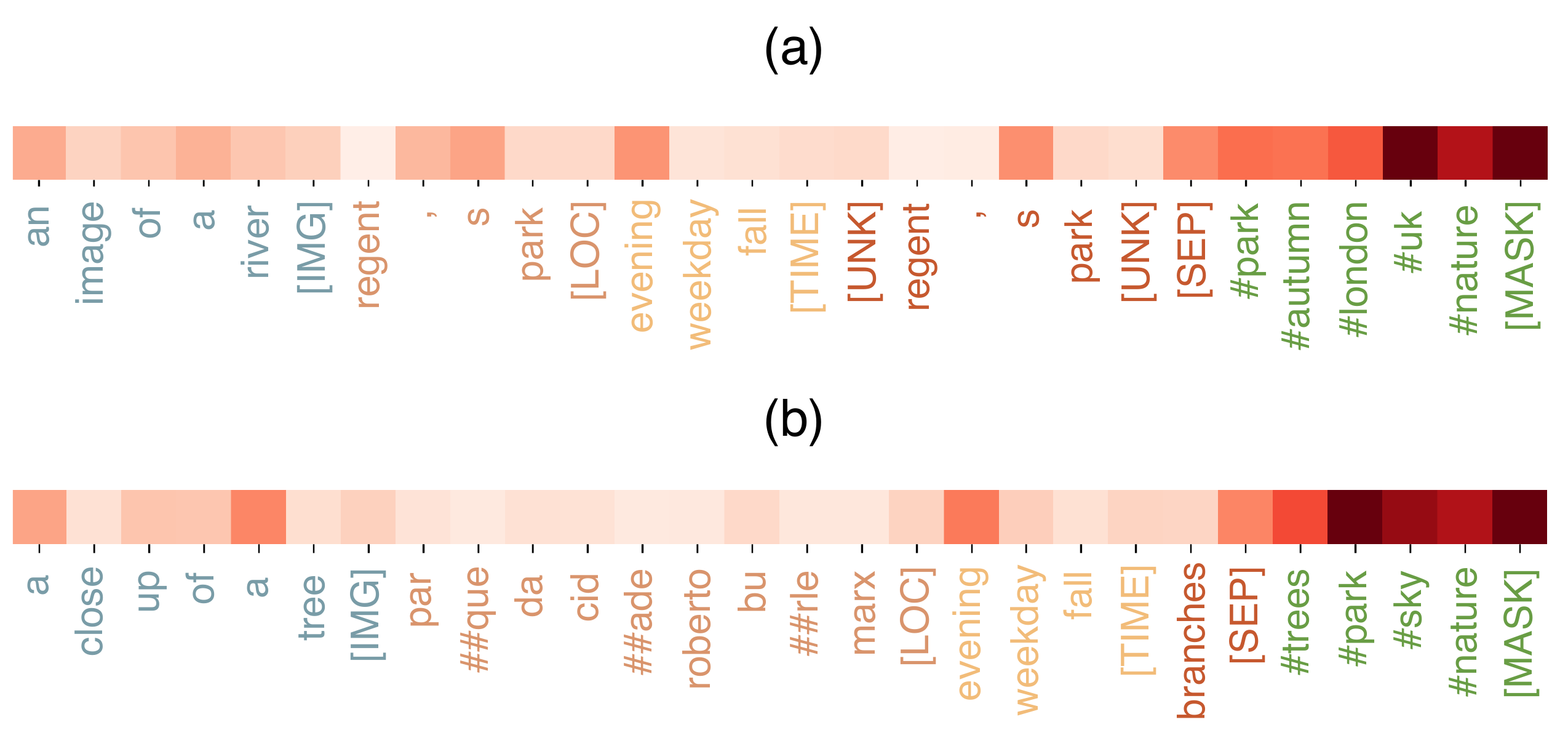}
    \caption{Attention scores with color intensity, showing how much [MASK] tokens attend to other tokens.}
    \label{fig:mask_attention}
\end{figure}

\begin{figure}
    \centering
    \includegraphics[width=0.48\textwidth]{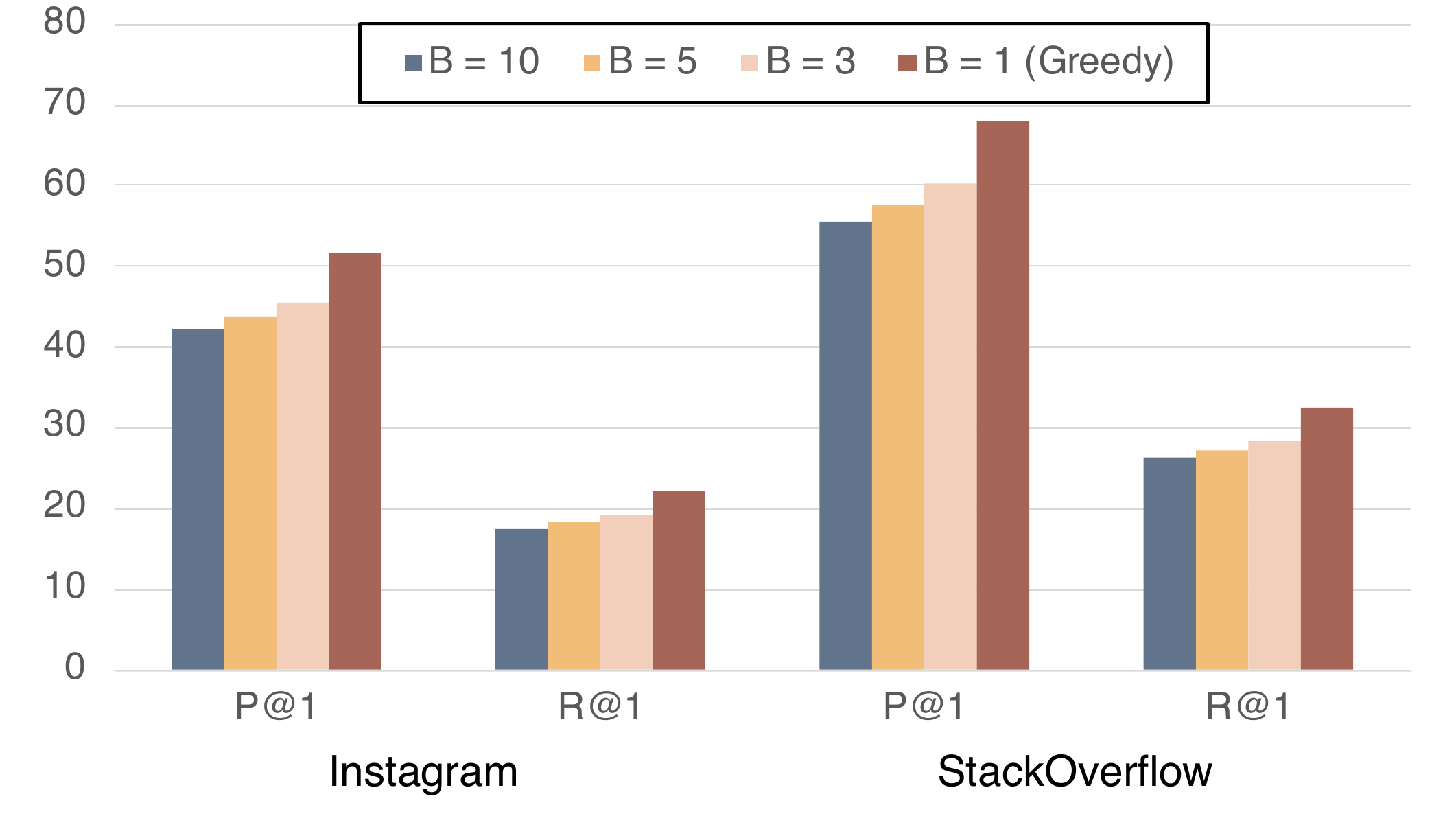}
    \caption{Sequence-oblivious decoding (B=1) and beam search results with different beam widths.}
    \label{fig:beam_search}
\end{figure}

\begin{figure*}[h!]
    \centering
    \includegraphics[width=0.95\textwidth]{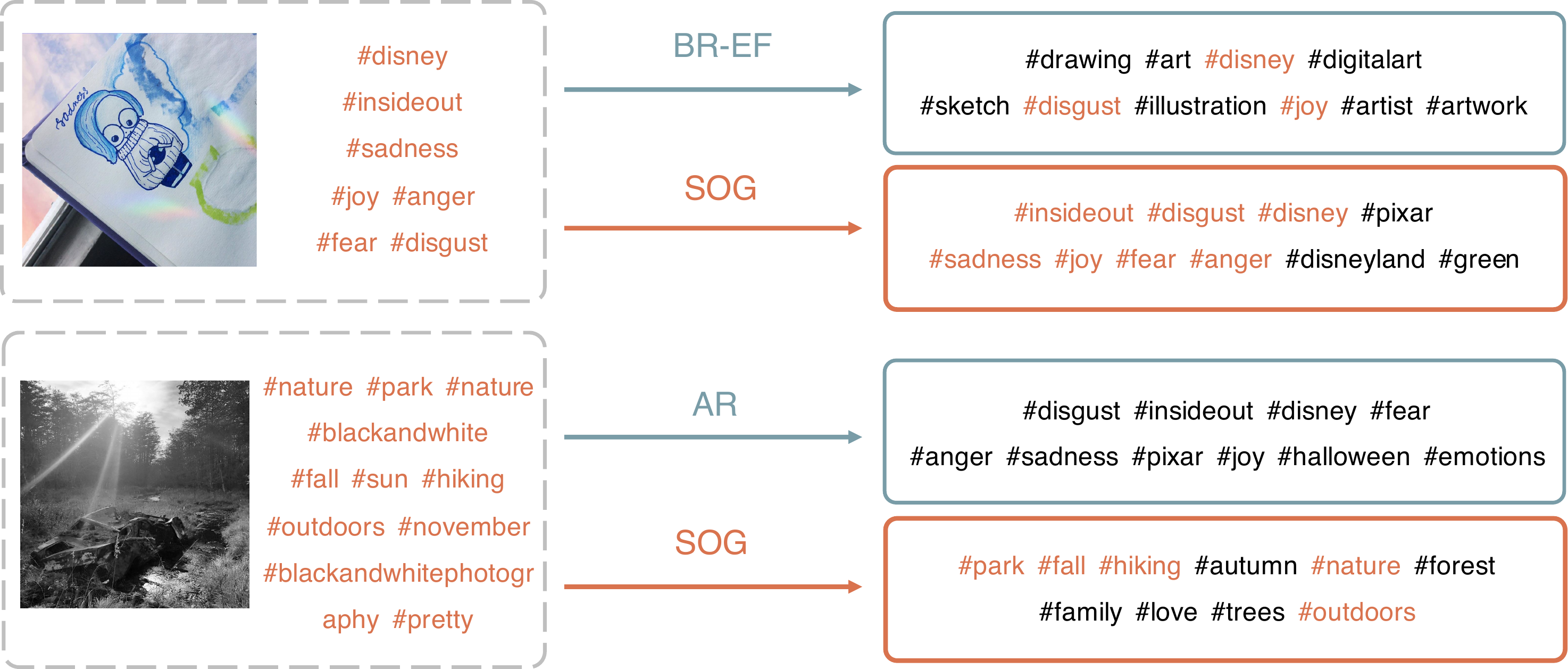}
    \caption{A qualitative analysis of tags generated by different models. The predicted tags in red are correct ones.}
    \label{fig:qualitative_analysis}
\end{figure*}

\subsection{On Inter-Dependency}
Unlike the AR approach of using the immediately preceding token, we use the \texttt{[MASK]} token to fully exploit the previously generated tag set. Figure \ref{fig:mask_attention} (a) and (b) illustrate how much the \texttt{[MASK]} token attends to the other tokens. Evidently, much stronger attention is paid to the generated tags than other tokens. This shows that the previously generated tags play a critical role in predicting the next tag, which is in accordance with our claim for the need to consider \textit{inter-dependency} among the target tags. This phenomenon is also apparent in Table \ref{table:interactions}, where it shows high average tag-to-tag interaction (63.31\%). It is worth noting that the \texttt{[MASK]} token is not heavily relying on the last tag.

\subsection{Greedy vs. Beam Search}
%To see if the commonly used beam search strategy in text generation is also applicable in tag recommendation, 
In order to ensure the validity of our sequence-oblivious decoding scheme, we compare the greedy and beam search strategies.
We apply beam search with different width settings ($B$ = {1, 3, 5, 10}) for different numbers of candidate sequences. Here, $B = 1$ is equal to greedy since it takes the most probable item at each time step. In Figure \ref{fig:beam_search}, we observe a substantial performance drop in our model when we applied beam search to tag generation. As mentioned in \ref{decoding}, this result can be explained in terms of the characteristic differences between a text sequence and a tag sequence, confirming greedy decoding is more natural and amenable to tag recommendation than beam search. %We validate our hypothesis that greedy decoding is naturally sequence-oblivious and more amenable to tag recommendation than beam search, which is the de facto standard in sequential text generation tasks.

\begin{table}[h]
    {\small
    \centering
    \begin{tabularx}{\linewidth}{ c *{6}{Y} }
    \toprule
        \  & Image & Location & Time & Text & Tag \\ \midrule\midrule
        Image & 39.74\% & 13.65\% & 22.33\% & 9.76\% & 14.52\% \\ 
        Location & 13.48\% & 47.01\% & 19.37\% & 8.80\% & 11.34\%  \\
        Time & 11.07\% & 11.61\% & 51.67\% & 10.77\% & 14.88\%  \\ 
        Text & 12.39\% & 10.37\% & 16.71\% & 39.19\% & 21.34\%  \\ 
        Tag & 9.29\% & 6.72\% & 10.27\% & 10.41\% & 63.31\%  \\ \midrule
        \textbf{[MASK]} & 14.27\% & 9.28\% & 11.58\% & 10.86\% & 54.01\%  \\ \bottomrule
    \end{tabularx}
    }
    \caption{Interaction among feature types. Rows show how much feature types attend to others. Feature-wise scores are normalized over all attention heads and layers and averaged over all instances in the test set.}
    \label{table:interactions}
\end{table}

\begingroup
\begin{table}[h!]
    {\small
    \centering
    \begin{tabularx}{\linewidth}{ c *{7}{Y} }
    \toprule
    \textbf{Condition} & \textbf{P@1} & \textbf{P@3} & \textbf{P@5} & \textbf{R@1} & \textbf{R@3} & \textbf{R@5} \\ \midrule\midrule
    \textbf{All Contexts}  &   \textbf{51.85}   &   \textbf{30.15}   &   \textbf{22.03}   &   \textbf{22.29}   &   \textbf{32.38}   &   \textbf{36.62}   \\ \midrule
    w/o Image\ \ \ \ \   &   40.57   &   21.69   &   15.72    &   17.44   &   23.50   &   26.29   \\ 
    w/o Location  &   49.79   &   28.51   &   20.85   &   21.29   &   30.61   &   34.64   \\ 
    w/o Time\ \ \ \ \ \ \   &   43.23   &   25.96   &   19.27   &   18.49   &   27.69   &   31.64   \\
    w/o Text\ \ \ \ \ \ \ \   &   37.24   &   20.48   &   14.74   &   15.73   &   21.82   &   24.35   \\ \bottomrule
    \end{tabularx}
    }
    \caption{Ablation over the feature types using our model. Only one feature is removed at a time.}
    \label{table:ablation}
\end{table}
\endgroup

\subsection{Impact of Context-Awareness}
We also conduct an ablation study to see how each feature type contributes to the model performance. In Table \ref{table:ablation}, every evaluation score decreases when we remove one of the input feature types, implying all of the feature types contribute to the model prediction. It shows that \textit{Text} is the most important, probably because it comes directly from the users and is most native to the BERT language model. On the other hand, the \textit{location} and \textit{time} features appear to be less important because they are secondary descriptions derived from the original descriptor. Usually the text form of location is too specific and diverse for the model to capture the patterns. Table \ref{table:interactions} also reveals the degree of interactions among the feature types. The attention scores indicate that the features of one type attend to other features (row-wise) quite actively, corroborating our assumption that the context features in a post are mutually complementary. 

%The distributed, scattered interaction among context features show their active interactions (i.e., trained attention weights), 
%This also explains the superiority of our early fusion approach to late fusion.

% \subsection{Attention Analysis}
% We examine the level of interactions among the feature types, the context-to-tag correlation, and the importance of generated tags as well as the feature types in the \texttt{[MASK]} token representation. Table \ref{table:interactions} shows the averaged attention scores for each feature over the test set, which indicate how much the feature attend to the other features (row-wise). It shows that the features interact with one another quite actively as expected. We note that a major portion of attention from \texttt{[MASK]} (54.01\%) is on the previous tags, witnessing the dependency among tags manifests itself in our model. In Figure \ref{fig:mask_attention} (a) and (b), which illustrate how much \texttt{[MASK]} token attends to other tokens, we observe that it largely focuses on the previous tags. This shows that the generated tags play a critical role in generating a related tag, which is in accordance with our claim.

\subsection{Qualitative Analysis}
Figure \ref{fig:qualitative_analysis} shows two test cases that compare outputs that different models generate for the posts. For the top post, the BR-EF model produces relevant tags like \textit{\#disney} but fails to predict others like \textit{\#fear} and \textit{\#anger} that our model generates successfully, which require inter-tag dependency (Joy, Fear and Anger are characters from the Disney movie, \textit{Inside Out}).
For the bottom post, the AR model fails to generate any of the gold tags, because it heavily relies on the immediately preceding tag instead of the entire context. Since the model initially produced wrong hashtags (\textit{\#disgust}, \textit{\#insideout}), the AR model propagates the erroneous tag information throughout the subsequent generations.
%In contrast, our model generates many of the correct tags successfully, showing its accountability for direct view of contextual information.

\section{Conclusion}
This paper characterizes the tag recommendation tasks with and without multi-modal context information in the posts of Instagram and Stack Overflow, respectively, and proposes a novel framework, \textit{sequence-oblivious} generation (SOG), that explicitly considers the inter-tag dependency and the orderless nature of tags. We address the drawbacks of the conventional ranking and AR approaches to tag recommendation and define it in a new way so that it attends to the characteristics of ``tag language." For the new framework, we design the sequence-oblivious model and training and decoding strategies, together with the BERT-based early fusion method for multi-modal features.
%To exploit the rich, multi-modal feature information, we adopt an early fusion approach by converting distinct features to the same modality (i.e., text) and leveraging BERT's language understanding capabilities. 
In the extensive experiments on two different domains, we show that SOG outperforms the the baselines by significant margins. Also shown are the roles of the iterative query expansion with generated tags, the 1-to-M training scheme under the \textit{orderlessness} assumption, the early fusion method over late fusion, and the adoption of greedy search for decoding. For future work, we plan to investigate if our generative framework can generalize over other tasks possessing the ``tag language" characteristics.

\section*{Acknowledgements}

This work was supported by Institute for Information \& communications Technology Planning \& Evaluation(IITP) grant funded by the Korea government(MSIT) (No. 2013-2-00131, Development of Knowledge Evolutionary WiseQA Platform Technology for Human Knowledge Augmented Services).

% This document has been adapted
% by Steven Bethard, Ryan Cotterell and Rui Yan
% from the instructions for earlier ACL and NAACL proceedings, including those for 
% ACL 2019 by Douwe Kiela and Ivan Vuli\'{c},
% NAACL 2019 by Stephanie Lukin and Alla Roskovskaya, 
% ACL 2018 by Shay Cohen, Kevin Gimpel, and Wei Lu, 
% NAACL 2018 by Margaret Mitchell and Stephanie Lukin,
% Bib\TeX{} suggestions for (NA)ACL 2017/2018 from Jason Eisner,
% ACL 2017 by Dan Gildea and Min-Yen Kan, 
% NAACL 2017 by Margaret Mitchell, 
% ACL 2012 by Maggie Li and Michael White, 
% ACL 2010 by Jing-Shin Chang and Philipp Koehn, 
% ACL 2008 by Johanna D. Moore, Simone Teufel, James Allan, and Sadaoki Furui, 
% ACL 2005 by Hwee Tou Ng and Kemal Oflazer, 
% ACL 2002 by Eugene Charniak and Dekang Lin, 
% and earlier ACL and EACL formats written by several people, including
% John Chen, Henry S. Thompson and Donald Walker.
% Additional elements were taken from the formatting instructions of the \emph{International Joint Conference on Artificial Intelligence} and the \emph{Conference on Computer Vision and Pattern Recognition}.

\bibliography{anthology,custom}

\begin{thebibliography}{34}
\expandafter\ifx\csname natexlab\endcsname\relax\def\natexlab#1{#1}\fi

\bibitem[{Bahdanau et~al.(2014)Bahdanau, Cho, and Bengio}]{bahdanau2014neural}
Dzmitry Bahdanau, Kyunghyun Cho, and Yoshua Bengio. 2014.
\newblock \href {http://arxiv.org/abs/1409.0473} {Neural machine translation by
  jointly learning to align and translate}.
\newblock Cite arxiv:1409.0473Comment: Accepted at ICLR 2015 as oral
  presentation.

\bibitem[{Chan and Fan(2019{\natexlab{a}})}]{chan-fan-2019-bert}
Ying-Hong Chan and Yao-Chung Fan. 2019{\natexlab{a}}.
\newblock \href {https://doi.org/10.18653/v1/W19-8624} {{BERT} for question
  generation}.
\newblock In \emph{Proceedings of the 12th International Conference on Natural
  Language Generation}, pages 173--177, Tokyo, Japan. Association for
  Computational Linguistics.

\bibitem[{Chan and Fan(2019{\natexlab{b}})}]{chan-fan-2019-recurrent}
Ying-Hong Chan and Yao-Chung Fan. 2019{\natexlab{b}}.
\newblock \href {https://doi.org/10.18653/v1/D19-5821} {A recurrent
  {BERT}-based model for question generation}.
\newblock In \emph{Proceedings of the 2nd Workshop on Machine Reading for
  Question Answering}, pages 154--162, Hong Kong, China. Association for
  Computational Linguistics.

\bibitem[{Chi et~al.(2020)Chi, Dong, Wei, Wang, Mao, and Huang}]{chi2020cross}
Zewen Chi, Li~Dong, Furu Wei, Wenhui Wang, Xian-Ling Mao, and Heyan Huang.
  2020.
\newblock Cross-lingual natural language generation via pre-training.
\newblock In \emph{Proceedings of the AAAI Conference on Artificial
  Intelligence}, volume~34, pages 7570--7577.

\bibitem[{Cho et~al.(2014)Cho, van Merri{\"e}nboer, Bahdanau, and
  Bengio}]{cho-etal-2014-properties}
Kyunghyun Cho, Bart van Merri{\"e}nboer, Dzmitry Bahdanau, and Yoshua Bengio.
  2014.
\newblock \href {https://doi.org/10.3115/v1/W14-4012} {On the properties of
  neural machine translation: Encoder{--}decoder approaches}.
\newblock In \emph{Proceedings of {SSST}-8, Eighth Workshop on Syntax,
  Semantics and Structure in Statistical Translation}, pages 103--111, Doha,
  Qatar. Association for Computational Linguistics.

\bibitem[{Clark et~al.(2020)Clark, Luong, Le, and Manning}]{Clark2020ELECTRA:}
Kevin Clark, Minh-Thang Luong, Quoc~V. Le, and Christopher~D. Manning. 2020.
\newblock \href {https://openreview.net/forum?id=r1xMH1BtvB} {Electra:
  Pre-training text encoders as discriminators rather than generators}.
\newblock In \emph{International Conference on Learning Representations}.

\bibitem[{Denton et~al.(2015)Denton, Weston, Paluri, Bourdev, and
  Fergus}]{facebook_hashtag_prediction}
Emily Denton, Jason Weston, Manohar Paluri, Lubomir Bourdev, and Rob Fergus.
  2015.
\newblock User conditional hashtag prediction for images.
\newblock In \emph{Proceedings of the 21th ACM SIGKDD international conference
  on knowledge discovery and data mining}, pages 1731--1740.

\bibitem[{Devlin et~al.(2019)Devlin, Chang, Lee, and
  Toutanova}]{Devlin2019BERTPO}
J.~Devlin, Ming-Wei Chang, Kenton Lee, and Kristina Toutanova. 2019.
\newblock Bert: Pre-training of deep bidirectional transformers for language
  understanding.
\newblock In \emph{NAACL-HLT}.

\bibitem[{Ding et~al.(2012)Ding, Zhang, and Huang}]{ding2012automatic}
Zhuoye Ding, Qi~Zhang, and Xuan-Jing Huang. 2012.
\newblock Automatic hashtag recommendation for microblogs using topic-specific
  translation model.
\newblock In \emph{Proceedings of COLING 2012: Posters}, pages 265--274.

\bibitem[{Godin et~al.(2013)Godin, Slavkovikj, De~Neve, Schrauwen, and Van~de
  Walle}]{godin2013using}
Fr{\'e}deric Godin, Viktor Slavkovikj, Wesley De~Neve, Benjamin Schrauwen, and
  Rik Van~de Walle. 2013.
\newblock Using topic models for twitter hashtag recommendation.
\newblock In \emph{Proceedings of the 22nd International Conference on World
  Wide Web}, pages 593--596.

\bibitem[{Gong and Zhang(2016)}]{gong2016hashtag}
Yuyun Gong and Qi~Zhang. 2016.
\newblock Hashtag recommendation using attention-based convolutional neural
  network.
\newblock In \emph{Proceedings of the Twenty-Fifth International Joint
  Conference on Artificial Intelligence}, IJCAI'16, page 2782–2788. AAAI
  Press.

\bibitem[{Kaviani and Rahmani(2020)}]{kaviani2020emhash}
Mohadeseh Kaviani and Hossein Rahmani. 2020.
\newblock Emhash: Hashtag recommendation using neural network based on bert
  embedding.
\newblock In \emph{2020 6th International Conference on Web Research (ICWR)},
  pages 113--118. IEEE.

\bibitem[{Lewis et~al.(2004)Lewis, Yang, Rose, and
  Li}]{10.5555/1005332.1005345}
David~D. Lewis, Yiming Yang, Tony~G. Rose, and Fan Li. 2004.
\newblock Rcv1: A new benchmark collection for text categorization research.
\newblock \emph{J. Mach. Learn. Res.}, 5:361–397.

\bibitem[{Li et~al.(2016)Li, Shah, Nourbakhsh, Liu, and Fang}]{li2016hashtag}
Quanzhi Li, Sameena Shah, Armineh Nourbakhsh, Xiaomo Liu, and Rui Fang. 2016.
\newblock Hashtag recommendation based on topic enhanced embedding, tweet
  entity data and learning to rank.
\newblock In \emph{Proceedings of the 25th ACM International on Conference on
  Information and Knowledge Management}, pages 2085--2088.

\bibitem[{Li et~al.(2019)Li, Liu, Hu, and Jiang}]{li2019topical}
Yang Li, Ting Liu, Jingwen Hu, and Jing Jiang. 2019.
\newblock Topical co-attention networks for hashtag recommendation on
  microblogs.
\newblock \emph{Neurocomputing}, 331:356--365.

\bibitem[{Park et~al.(2016)Park, Li, and Kim}]{park2016harrison}
Minseok Park, Hanxiang Li, and Junmo Kim. 2016.
\newblock Harrison: A benchmark on hashtag recommendation for real-world images
  in social networks.
\newblock \emph{arXiv preprint arXiv:1605.05054}.

\bibitem[{Radford et~al.(2018)Radford, Narasimhan, Salimans, and
  Sutskever}]{radford2018improving}
Alec Radford, Karthik Narasimhan, Tim Salimans, and Ilya Sutskever. 2018.
\newblock Improving language understanding by generative pre-training.

\bibitem[{Sedhai and Sun(2014)}]{sedhai2014hashtag}
Surendra Sedhai and Aixin Sun. 2014.
\newblock Hashtag recommendation for hyperlinked tweets.
\newblock In \emph{Proceedings of the 37th international ACM SIGIR conference
  on Research \& development in information retrieval}, pages 831--834.

\bibitem[{Sutskever et~al.(2014)Sutskever, Vinyals, and
  Le}]{sutskever2014sequence}
Ilya Sutskever, Oriol Vinyals, and Quoc~V Le. 2014.
\newblock Sequence to sequence learning with neural networks.
\newblock In \emph{Advances in neural information processing systems}, pages
  3104--3112.

\bibitem[{Vaswani et~al.(2017)Vaswani, Shazeer, Parmar, Uszkoreit, Jones,
  Gomez, Kaiser, and Polosukhin}]{vaswani2017attention}
Ashish Vaswani, Noam Shazeer, Niki Parmar, Jakob Uszkoreit, Llion Jones,
  Aidan~N Gomez, {\L}ukasz Kaiser, and Illia Polosukhin. 2017.
\newblock Attention is all you need.
\newblock In \emph{Advances in neural information processing systems}, pages
  5998--6008.

\bibitem[{Wang et~al.(2019)Wang, Li, King, Lyu, and Shi}]{wang2019microblog}
Yue Wang, Jing Li, Irwin King, Michael~R Lyu, and Shuming Shi. 2019.
\newblock Microblog hashtag generation via encoding conversation contexts.
\newblock In \emph{Proceedings of the 2019 Conference of the North American
  Chapter of the Association for Computational Linguistics: Human Language
  Technologies, Volume 1 (Long and Short Papers)}, pages 1624--1633.

\bibitem[{Weston et~al.(2014)Weston, Chopra, and
  Adams}]{weston-etal-2014-tagspace}
Jason Weston, Sumit Chopra, and Keith Adams. 2014.
\newblock \href {https://doi.org/10.3115/v1/D14-1194} {{\#}{T}ag{S}pace:
  Semantic embeddings from hashtags}.
\newblock In \emph{Proceedings of the 2014 Conference on Empirical Methods in
  Natural Language Processing ({EMNLP})}, pages 1822--1827, Doha, Qatar.
  Association for Computational Linguistics.

\bibitem[{Wolf et~al.(2019)Wolf, Debut, Sanh, Chaumond, Delangue, Moi, Cistac,
  Rault, Louf, Funtowicz et~al.}]{wolf2019transformers}
Thomas Wolf, Lysandre Debut, Victor Sanh, Julien Chaumond, Clement Delangue,
  Anthony Moi, Pierric Cistac, Tim Rault, R{\'e}mi Louf, Morgan Funtowicz,
  et~al. 2019.
\newblock Transformers: State-of-the-art natural language processing.
\newblock \emph{arXiv preprint arXiv:1910.03771}.

\bibitem[{Wu et~al.(2018{\natexlab{a}})Wu, Li, Yan, Li, Gu, and
  Yang}]{wu2018hashtag}
Gaosheng Wu, Yuhua Li, Wenjin Yan, Ruixuan Li, Xiwu Gu, and Qi~Yang.
  2018{\natexlab{a}}.
\newblock Hashtag recommendation with attention-based neural image hashtagging
  network.
\newblock In \emph{International Conference on Neural Information Processing},
  pages 52--63. Springer.

\bibitem[{Wu et~al.(2018{\natexlab{b}})Wu, Fisch, Chopra, Adams, Bordes, and
  Weston}]{Wu2018StarSpaceEA}
Ledell~Yu Wu, Adam Fisch, Sumit Chopra, Keith Adams, Antoine Bordes, and Jason
  Weston. 2018{\natexlab{b}}.
\newblock Starspace: Embed all the things!
\newblock In \emph{AAAI}.

\bibitem[{Xu and Croft(1996)}]{10.1145/243199.243202}
Jinxi Xu and W.~Bruce Croft. 1996.
\newblock \href {https://doi.org/10.1145/243199.243202} {Query expansion using
  local and global document analysis}.
\newblock In \emph{Proceedings of the 19th Annual International ACM SIGIR
  Conference on Research and Development in Information Retrieval}, SIGIR '96,
  page 4–11, New York, NY, USA. Association for Computing Machinery.

\bibitem[{Yang et~al.(2020{\natexlab{a}})Yang, Wang, and
  Jiang}]{yang2020sentiment}
Chao Yang, Xiaochan Wang, and Bin Jiang. 2020{\natexlab{a}}.
\newblock Sentiment enhanced multi-modal hashtag recommendation for
  micro-videos.
\newblock \emph{IEEE Access}, 8:78252--78264.

\bibitem[{Yang et~al.(2019{\natexlab{a}})Yang, Luo, Ma, Lin, and
  Sun}]{yang-etal-2019-deep}
Pengcheng Yang, Fuli Luo, Shuming Ma, Junyang Lin, and Xu~Sun.
  2019{\natexlab{a}}.
\newblock \href {https://doi.org/10.18653/v1/P19-1518} {A deep reinforced
  sequence-to-set model for multi-label classification}.
\newblock In \emph{Proceedings of the 57th Annual Meeting of the Association
  for Computational Linguistics}, pages 5252--5258, Florence, Italy.
  Association for Computational Linguistics.

\bibitem[{Yang et~al.(2018)Yang, Sun, Li, Ma, Wu, and
  Wang}]{yang-etal-2018-sgm}
Pengcheng Yang, Xu~Sun, Wei Li, Shuming Ma, Wei Wu, and Houfeng Wang. 2018.
\newblock \href {https://www.aclweb.org/anthology/C18-1330} {{SGM}: Sequence
  generation model for multi-label classification}.
\newblock In \emph{Proceedings of the 27th International Conference on
  Computational Linguistics}, pages 3915--3926, Santa Fe, New Mexico, USA.
  Association for Computational Linguistics.

\bibitem[{Yang et~al.(2020{\natexlab{b}})Yang, Wu, Li, Li, Gu, Deng, and
  Wu}]{yang2020amnn}
Qi~Yang, Gaosheng Wu, Yuhua Li, Ruixuan Li, Xiwu Gu, Huicai Deng, and Junzhuang
  Wu. 2020{\natexlab{b}}.
\newblock Amnn: Attention-based multimodal neural network model for hashtag
  recommendation.
\newblock \emph{IEEE Transactions on Computational Social Systems}.

\bibitem[{Yang et~al.(2019{\natexlab{b}})Yang, Dai, Yang, Carbonell,
  Salakhutdinov, and Le}]{yang2019xlnet}
Zhilin Yang, Zihang Dai, Yiming Yang, Jaime Carbonell, Ruslan Salakhutdinov,
  and Quoc~V Le. 2019{\natexlab{b}}.
\newblock Xlnet: Generalized autoregressive pretraining for language
  understanding.
\newblock \emph{arXiv preprint arXiv:1906.08237}.

\bibitem[{Zangerle et~al.(2011)Zangerle, Gassler, and
  Specht}]{zangerle2011recommending}
E.~Zangerle, W.~Gassler, and G.~Specht. 2011.
\newblock Recommending \#-tags in twitter.
\newblock In \emph{Proceedings of the Workshop on Semantic Adaptive Social Web
  (SASWeb 2011). CEUR Workshop Proceedings}, volume 730, pages 67--78.

\bibitem[{Zhang et~al.(2019)Zhang, Yao, Xu, Tong, Yan, and
  Lu}]{zhang2019hashtag}
Suwei Zhang, Yuan Yao, Feng Xu, Hanghang Tong, Xiaohui Yan, and Jian Lu. 2019.
\newblock Hashtag recommendation for photo sharing services.
\newblock In \emph{Proceedings of the AAAI Conference on Artificial
  Intelligence}, volume~33, pages 5805--5812.

\bibitem[{Zhao et~al.(2016)Zhao, Zhu, Jin, and Yang}]{zhao2016personalized}
Feng Zhao, Yajun Zhu, Hai Jin, and Laurence~T Yang. 2016.
\newblock A personalized hashtag recommendation approach using lda-based topic
  model in microblog environment.
\newblock \emph{Future Generation Computer Systems}, 65(C):196--206.

\end{thebibliography}
\bibliographystyle{acl_natbib}

\clearpage

\appendix

\section{Appendix}
\label{sec:appendix}

\begin{table*}[ht!]
    \begin{center}
    \begin{tabular*}{0.7\linewidth}{cccccc}
    \toprule
    Hyperparameter & Joint Space & BR & Seq2Set & AR & SOG \\ \midrule\midrule
    Batch size   &   128   &   64   &   64   &   64   &   64   \\
    Epoch   &   10   &   5   &   20+20*   &   5 / 4   &   5   \\
    Hidden size   &   128   &   768   &   512   &   768   &   768   \\
    Maximum length   &   100 / 20   &   384   &   500   &   384   &   384   \\
    Learning rate   &   1e-3 / 1e-4   &   5e-5   &   3e-4   &   5e-5   &   5e-5   \\
    Optimizer   &   Adam   &   Adam   &   Adam   &   Adam   &   Adam   \\
    Seed   &   -   &   42   &   1234   &   42   &   42   \\
    Approx. runtime  &  0.2 hrs  &  7 hrs  &  10 hrs  &  7 hrs  &  25 hrs    \\
    Parameters &  1.3M &  110M  &  20M  &  140M  &  110M   \\
    \bottomrule
    \end{tabular*}
    \end{center}
    \caption{Hyperparameters of the models. Most settings are the same for both datasets. ``/" is used only when the setting differ between Instagram / Stack Overflow, due to the unusual case where they couldn't be trained well. * indicates 20 epochs for pre-training with MLE and another 20 epochs for fine-tuning with RL.}
    \label{table:hyperparameter}
\end{table*}

\subsection{Implementation Details of Baselines} \label{hyperparameters}
Here we provide the implementation details including hyperparameters (in Table \ref{table:hyperparameter}) and model architectures of Joint Space, Seq2Set and BERT-based (including BERT-based Ranking, AR, and SOG). The hyperparameter values have been manually tuned using the P@K and R@K criteria to select the best performing version for each model in our work. We checked the validity of the reported results by having 3 times of trials for each setting. The implementation details are as follows:
\begin{itemize}
    \item \textbf{Joint Space (Ranking).}
    We use a 3-layer and 2-layer LSTM to model the text description for Instagram and Stack Overflow, respectively, and a 10-layer convolutional layer with the kernel size equal to 3 for image encoding. For location and time, we use the representations from the context embedding matrix. If location and time consist of multiple tokens, we take an average over the token embeddings. For pairwise ranking loss, we randomly sample two negative tags from the entire tag set while excluding the ground truth tags.
    
    We have also experimented with other settings to find the best hyperparameters for the Joint Space model as follows:
    \begin{itemize}
        \item \textbf{Text Encoding:} 1D convolutional layer with the kernel size equal to 3 as in image encoding.
        \item \textbf{Image Encoding:} 1D convolutional layers with the kernel sizes equal to 3, 4 and 5, and aggregation of these three settings, respectively.
    \end{itemize}
    For a fair comparison between Joint Space and BERT-based Ranking, we replaced the encoding module of Joint Space with BERT to use the early fusion proposed in this work. We conducted several experiments with different embedding sizes (e.g., 128 and 768), only to find out that none of them were able to improve over the Joint Space for some reason and even worse than the Frequency-Based model, which we did not mention in this paper but generated tags purely based on their frequency.
    % 여러 실험들을 했는데 다 언급할 예정
    % 1. Embedding size 50
    % 2. 3, 4, 5-gram
    % 3. BERT 사용해서 Context encoding, but embedding size 768
    % 4. embedding size 128
    % Best performance를 achieve하는 Context-Tag Mapping Model을 Find하기 위해 여러 실험을 진행했다. Text encoding을 위해 Image와 같이 1d convolutional layer를 사용했고, Image encoding을 위해 kernel size가 3, 4, 5인 1D convolutional layer를 사용해서 이 output들을 aggregate했다. 
    % 또한 BERT-based Ranking model과 Approach 측면에서 비교를 하기 위해 BERT를 사용해서 context를 encoding 한 뒤 Context-Tag Mapping 방식을 통해 tag를 generate했다. context의 embedding size를 768, 128로 모두 실험을 해보았지만 frequency baselines보다 성능이 나오지 않았다.
    
    \item \textbf{BERT-based Ranking (LF).} For BERT-based Ranking with late fusion, we input each feature separately to the same model for independent encoding and aggregate them in the end. For a fair comparison with the early fusion model, we assign unique index to each context feature type (image, location, time, and text) before giving it as input to our model. For example, we assign a starting index $0$ for image features and $20$ for location features to avoid overlap. Such index allocation allows us to prevent underfitting in the modeling of each feature and relieve the burden of having to encode input of differing modalities with a limited range of parameters. To be more specific, since we are feeding in each context feature separately according to their modality, we preclude the possibility of underfitted parameters by feeding every context input into the same index position.

    \item \textbf{Seq2Set.} For implementation and evaluation, we follow most of the settings in \cite{yang-etal-2019-deep} the same. As mentioned in the official code and the paper, we pre-train the Seq2Seq model with MLE and fine-tune it with the proposed RL scheme, where we train 20 epochs for each phase. We use a learning rate of 3e-4 to make it converge on our dataset, which is slightly higher than the default setting in the official code. To be fair with our method, a greedy search is used at decoding step.

    \item \textbf{AR.} We train AR using the 5 epochs, which is the same with BR and SOG, for Instagram. In the case of Stack Overflow, however, we needed to use 4 epochs exceptionally as we observed it begins to perform worse from 5 epochs. To discover the optimal number of layers and give more chance for context-tag interaction as in the recurrent BERT model, we tested a 12-layer Transformer decoder for experiments. However, there was no performance gain even with a significant increase in training time. Therefore, we decided to use a one-layer Transformer decoder.
\end{itemize}

\subsection{Data Construction} \label{data_construction}
Here we describe the procedure how we constructed the two datasets for extensive and meaningful experiments.

\begingroup
\setlength{\tabcolsep}{16pt}
\begin{table*}[ht!]
    \begin{center}
    \begin{tabular*}{\linewidth}{cccccc}
    \toprule
    \textbf{Activity} & \textbf{Emotion} & \textbf{Event} & \textbf{Location} & \textbf{Object} & \textbf{Time} \\ \midrule
    driving & angry & anniversary & bar & bike & afternoon \\
    eating & depression & birthday & beach & book & evening \\
    golf & disgust & bridalshower & cafe & car & fall \\
    party & fear & competition & city & coffee & holiday \\
    reading & happy & farewell & gym & flower & morning \\
    running & hope & graduation & home & medal & night \\
    shopping & love & newyear & office & pizza & spring \\
    sleeping & pity & seminar & outdoors & shoes & summer \\
    swimming & sad & wedding & park & sky & weekend \\
    travel & worry & welcome & school & tree & winter \\ \bottomrule
    \end{tabular*}
    \end{center}
    \caption{Seed tags used for collecting Instagram posts.}
    \label{table:seed_tag}
\end{table*}
\endgroup

\textbf{Instagram}
To collect meaningful and diverse tags from Instagram, we first define a set of \textit{seed tags} based on the level of generality and frequency, as in Table \ref{table:seed_tag}. Seed tags consist of 6 general categories (\texttt{Activity}, \texttt{Emotion}, \texttt{Event}, \texttt{Location}, \texttt{Object}, \texttt{Time}), and each category constitutes 10 tags. For example, ``\textit{\#beach}" is assigned to \texttt{Location} and ``\textit{\#happy}" to \texttt{Emotion}. Using the seed tags we collect 180K posts from Instagram and filter out those with more than 20 tags, resulting in 87,872 posts and 190K unique tags. This filtering strategy is based on the rationale that the posts with exceptionally many tags are highly likely to be an advertisement. Moreover, \cite{park2016harrison}\footnote{This dataset contains images with co-referenced tags, but not contextual information required for our method. Thus, we construct our own dataset.} observed the top 1,000 out of 165K unique hashtags cover more than half of the total, meaning that most are too specific or unused. Based on this, we decided to filter out such meaningless hashtags. They are not likely to have discriminative power for search/recommendation. We filter out those with less than 400 frequency, resulting in the final set of 907 tags.

\textbf{Stack Overflow}
We utilize a part of a corpus, StackSample\footnote{https://www.kaggle.com/stackoverflow/stacksample}, 10\% of Stack Overflow Q\&A posts that is publicly available on Kaggle. It contains 1,200K questions (i.e., posts) with the corresponding list of answers and tags. As part of an effort to construct a quality dataset, we filter out questions where scores (i.e., reputation) are less than 5, resulting in 81,320 posts and 19K unique tags. For the target tag set, we end up with 3,897 tags by including only those with a minimum frequency of 10.

\end{document}